\documentclass[10pt,twocolumn,letterpaper]{article}

\usepackage{cvpr}              
\usepackage{caption}
\definecolor{cvprblue}{rgb}{0.21,0.49,0.74}
\usepackage[pagebackref,breaklinks,colorlinks,allcolors=cvprblue]{hyperref}

\usepackage{booktabs,multirow,makecell,adjustbox,array,xcolor,graphicx}
\usepackage[table]{xcolor}
\usepackage{marvosym}

\definecolor{GainGreen}{RGB}{0,128,0}
\definecolor{DropRed}{RGB}{180,30,30}
\definecolor{RowGray}{RGB}{235,235,235}
\newcommand{\partialrowcolor}{\cellcolor{RowGray}}

\newcommand{\gsub}[1]{\ensuremath{_{\textcolor{GainGreen}{\scriptsize \textbf{#1}}}}} 
\newcommand{\dsub}[1]{\ensuremath{_{\textcolor{DropRed}{\scriptsize \textbf{#1}}}}}   
\newcommand{\odaft}{\textbf{ORION}}

\def\paperID{xxxx} 
\def\confName{CVPR}
\def\confYear{2026}

\title{\odaft{}: ORthonormal Text Encoding for Universal VLM AdaptatION}

\author{Omprakash Chakraborty\textsuperscript{\Letter} \and Jose Dolz \and Ismail Ben Ayed
\\
\and ÉTS Montréal, Canada
\\
{\textsuperscript{\Letter}  \tt\small omprakash.chakraborty@etsmtl.ca}
}

\begin{document}
\maketitle
\begin{abstract}

Vision-language models (VLMs) have demonstrated remarkable generalization across diverse tasks, yet their performance remains constrained by the quality and geometry of the textual prototypes used to represent classes. Standard zero-shot classifiers, derived from frozen text encoders and handcrafted prompts, may yield correlated or weakly separated embeddings that limit task-specific discriminability. We introduce \odaft{}, a text encoder fine-tuning framework that improves pretrained VLMs using only class names. Our method optimizes, via low-rank adaptation, a novel loss integrating two terms, one promoting pairwise orthogonality between the textual representations of the classes of a given task and the other penalizing deviations from the initial class prototypes. Furthermore,  we provide a probabilistic interpretation of our orthogonality penalty, connecting it to the general maximum likelihood estimation (MLE) principle via Huygens’ theorem. We report extensive experiments on 11 benchmarks and three large VLM backbones, showing that the refined textual embeddings yield powerful replacements for the standard CLIP prototypes. Added as {\em plug-and-play} module on top of various state-of-the-art methods, and across different prediction settings (zero-shot, few-shot and test-time adaptation), \odaft{} improves the performance consistently and significantly. We make the code available at \href{https://github.com/omchakrabarty/ORION.git}{\texttt{https://github.com/ORION}}.

\end{abstract}
    
\section{Introduction}
\label{sec:intro}

\begin{figure}[t]
    \centering
    \includegraphics[width=1.2\linewidth]{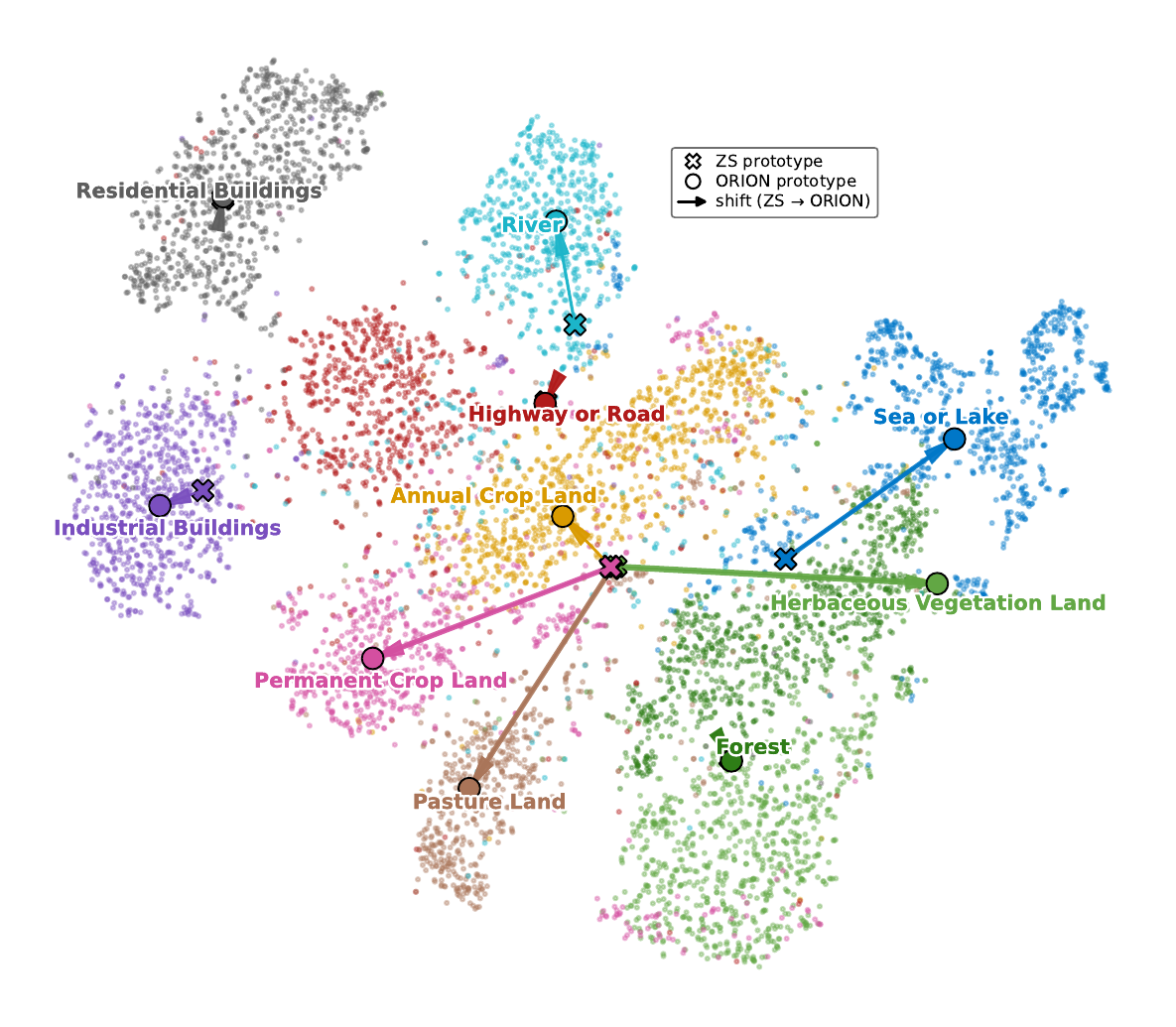}
    \vspace{-0.3em}
    \caption{
    \textbf{Motivation of \odaft{}.} Visualization on EuroSAT~\cite{helber2019eurosat} shows that semantically related categories such as \textit{Crop Land} and \textit{Pasture Land} are highly entangled in CLIP’s zero-shot space, with their textual prototypes ($\times$) misaligned from visual clusters. Our orthonormal text encoder (\odaft{}) re-centers them ($\circ$) toward the true image manifolds, improving class separation and inter-class geometry. Best viewed in color.
  }
    \label{fig:tsne_eurosat}
    \vspace{-0.7em}
\end{figure}

\begin{figure*}[t]
    \centering
    \includegraphics[width=0.9\textwidth]{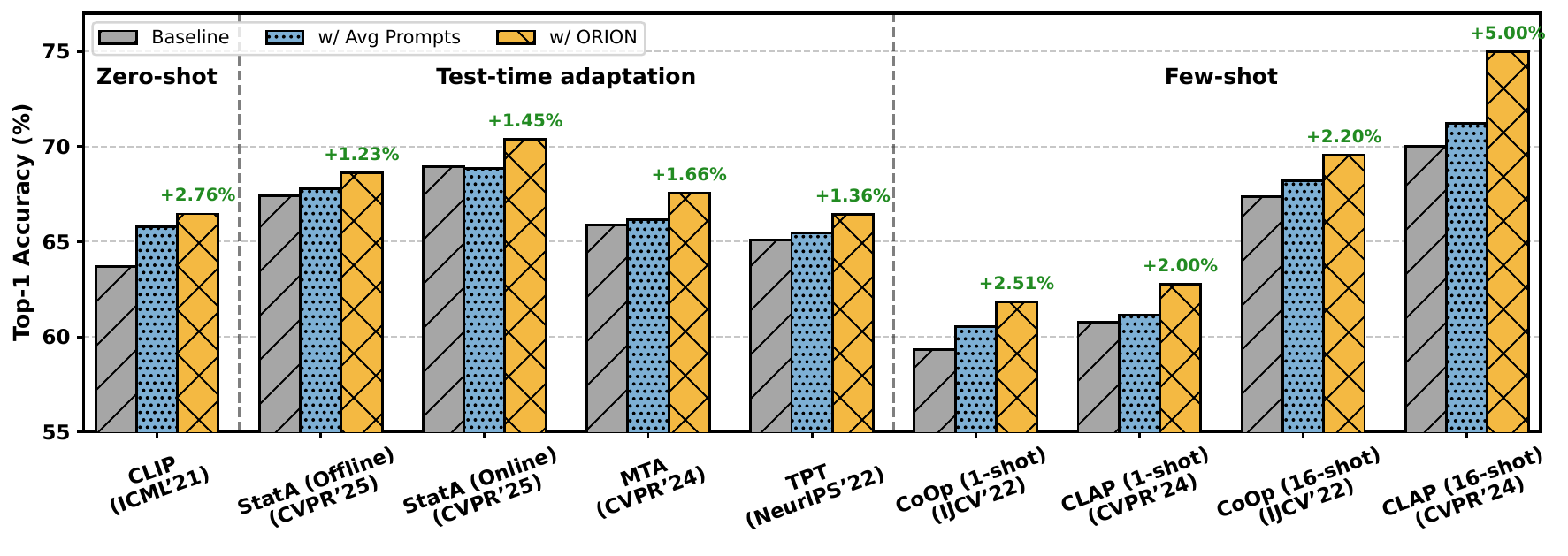}
    \vspace{-2mm}
    \caption{\textbf{Motivational overview.} Our text-only orthogonal fine-tuning (\textbf{\odaft}) uses \emph{only class names}—no images or captions—to refine the textual encoder of a frozen VLM.  
Across \textbf{zero-shot} (CLIP), \textbf{test-time adaptation} (StatA offline/online, MTA, TPT), and \textbf{few-shot} (CoOp/CLAP, 1- and 16-shot) settings, \odaft{} consistently improves Top-1 accuracy (averaged over 11 datasets).  
Bars correspond to \emph{Baseline}, \emph{+Avg Prompts}, and \emph{\odaft}; numbers above the bars (in \textcolor{GainGreen}{\textbf{green}}) indicate absolute gains over the respective baselines.  
By optimizing only the textual prototypes, \odaft{} yields a universal classifier that enhances diverse adaptation regimes without any visual supervision.}
    \label{fig:odafT_overview}
    \vspace{-3mm}
\end{figure*}

Vision--language models (VLMs) such as CLIP~\cite{radford2021clip}, MetaCLIP~\cite{xu2023metaclip} and ALIGN~\cite{jia2021align} 
have revolutionized multimodal learning by aligning vision and language in a shared embedding space.
Through large-scale contrastive pretraining, these models enable powerful zero-shot transfer across diverse visual tasks.
However, adapting pretrained VLMs to new datasets remains challenging: 
The representations of class names used as textual prototypes are often suboptimal for downstream domains.  
As a result, a growing body of work in the literature has focused on prompt tuning~\cite{zhou2022coop,zhou2022cocoop} and visual adaptation~\cite{shu2022tpt, zanella2024mta}, aiming to bridge the domain gap while keeping the base model frozen.  
Yet, the role of the text encoder, which defines the classifier space of a VLM, still remains greatly unexplored.  
This overlooked component is central to how the decision boundaries are formed in the shared embedding space and, as we show in this work, its geometry may influence significantly the generalization behavior of VLMs across zero-shot, few-shot and test-time adaptation settings.

The performance of VLMs is often bounded by the quality and geometry of the textual prototypes used for classification.
Existing works mainly refine prompts~\cite{zhou2022coop,zhou2022cocoop} or adapt the visual encoder~\cite{shu2022tpt,niu2023eta}, while the textual encoder typically remains frozen.
In standard zero-shot or few-shot evaluation, the performance of VLMs is highly sensitive to the phrasing of textual prompts---for example, ``\texttt{a photo of a \{class\}}'' versus ``\texttt{this is an image of a \{class\}.}''  
To reduce this prompt sensitivity, a common practice~\cite{osowiechi2024watt, stata2025cvpr, allingham2023simple, lu2022prompt, clap24, li2025modeling} is to average the embeddings of several text prompt templates per class, effectively smoothing over linguistic variations.  
While such averaging stabilizes performance, it may reduce the semantic diversity of class representations: The resulting prototypes may become correlated and occupy a narrow subspace of the textual embedding manifold.
Consequently, inter-class distinctions in the textual space may be weakened, limiting the discriminative power of the classifier.
Figure~\ref{fig:tsne_eurosat} provides an illustration of this using the EuroSAT~\cite{helber2019eurosat} fine-grained classification dataset.
In CLIP’s zero-shot space, semantically similar categories, such as \textit{Crop Land}, \textit{Pasture Land} and \textit{Herbaceous Vegetation Land}, have almost indistinguishable zero-shot prototypes; see the colored crosses in the Figure.
Being trained on general tasks, the initial zero-shot encoder does not discriminate between these fine-grained types of land, treating them as a general ``Land'' category. 
After deploying our orthonormal text encoder fine-tuning (\odaft), these category prototypes become well-separated and consistent with the corresponding visual data; see the colored circles in Figure~\ref{fig:tsne_eurosat}. This reveals that, even without the vision features, fine-tuning the text space alone can substantially improve class discriminability.
Interestingly, the separation is also \emph{task-adaptive}: \odaft{} enforces stronger repulsion among highly confounded classes while keeping already distinct ones (e.g., \textit{Crop Land} vs. \textit{Residential Buildings}) relatively stable. 
Such behavior reveals that the proposed soft orthonormality penalty does not impose uniform angular spacing; rather, it preserves the 
topology of the embedding space, enhancing discriminability precisely where fine-grained confusion occurs.

These observations raise a fundamental question: \emph{Can we improve the discriminative power of the text encoder of a VLM using only the class names as task-specific information, without relying on images?} Specifically, we propose to fine-tune the text encoder by minimizing a Frobenius-norm penalty, which promotes pairwise orthogonality constraints, encouraging angular diversity and enhancing the discriminative power of the textual class representations. 
As depicted in Figure~\ref{fig:odafT_overview}, our orthogonal refinement consistently delivers significant improvements across zero-shot, few-shot, and test-time adaptation settings.
The ensuing classifiers, learned without images, could serve as universal plug-and-play replacements for standard CLIP prototypes, providing a simple yet powerful way to enhance VLMs through geometric regularization of the textual space.

\noindent The key contributions are summarized as follows:
\begin{itemize}
    \item We introduce \odaft{}, a text encoder fine-tuning framework that improves pretrained VLMs using only class names. Our method follows from optimizing, via low-rank adaptation (LoRA), a novel composite loss containing two terms: i) a Frobenius-norm penalty promoting pairwise orthogonality between the textual representations of the classes of a given task; and ii) a term penalizing deviations from the initial class prototypes.  

    \item We provide a probabilistic interpretation of our orthogonality penalty, connecting it to the general maximum likelihood estimation (MLE) principle via Huygens’ theorem.
    
   \item We report extensive experiments on 11 benchmarks and three large backbones (CLIP ViT-B/16, CLIP ViT-L/14, and MetaCLIP), showing that the refined textual embeddings from \odaft{} act as powerful replacements for the standard CLIP prototypes. Added as plug-and-play module on top of various state-of-the-art methods, and across different prediction settings (\emph{zero-shot}, \emph{few-shot}, and \emph{test-time adaptation}), \odaft{} improves the performance consistently and significantly; see Figure ~\ref{fig:odafT_overview}. 
   
\end{itemize}

\section{Related Work}
\label{sec:related}

\noindent\textbf{Vision–Language Models and Few-shot Learning.}
Large vision–language models (VLMs) such as CLIP~\cite{radford2021clip}, MetaCLIP~\cite{xu2023metaclip}, and ALIGN~\cite{jia2021align} have demonstrated strong zero-shot recognition by aligning image and text modalities in a shared embedding space. 
Their zero-shot transferability relies critically on how class names are represented in the textual branch. 
To further improve downstream specialization, recent work explores few-shot learning, where trainable textual tokens or lightweight adapters are optimized while keeping the vision backbone frozen. 
Representative methods include CoOp~\cite{zhou2022coop} and CoCoOp~\cite{zhou2022cocoop}, which learn task-specific context tokens; TIP-Adapter and TIP-Adapter-F~\cite{zhang2022tipadapter}, which introduce fast key–value adaptation; and CLAP~\cite{clap24}, which refines the multimodal alignment through lightweight adapters rather than prompt optimization. 
TransCLIP~\cite{zanella2024boosting} and PLOT~\cite{plot2023} extend this idea toward transferable and cross-dataset prompts. 
Despite these advances, most approaches modify prompts 
or adapters, leaving the underlying geometry of the text encoder unchanged. 
Our work revisits this textual space itself, showing that enforcing orthogonality among class embeddings produce more discriminative and generalizable zero-shot representations that transfer seamlessly to few-shot settings. We focus our integration and analysis on CoOp and CLAP because they represent two canonical families of few-shot learning: (i) context-based prompt learning (CoOp) and (ii) adapter-based tuning (CLAP), both widely adopted and extensible to downstream few-shot frameworks. 

\noindent\textbf{Test-Time Adaptation for Vision–Language Models.}
To mitigate the degradation of pretrained models under distribution shift, recent works explore \emph{test-time adaptation} (TTA) strategies that adjust feature statistics or predictions at inference. 
TENT~\cite{wang2021tent}, SAR~\cite{niu2023eta} and CoTTA~\cite{wang2022cotta} adapt model normalization or entropy objectives without labels, while TPT~\cite{shu2022tpt} extends this principle to CLIP by refining pseudo-labels through prompt ensembles. 
Subsequent advances such as MTA~\cite{zanella2024mta} introduce multi-teacher averaging to stabilize predictions, and StatA~\cite{stata2025cvpr} proposes a realistic benchmark simulating non-i.i.d.\ and temporally correlated test streams. 
However, these approaches predominantly modify the visual or joint embeddings during adaptation. 
In contrast, \odaft{} improves robustness from the text side: by orthogonalizing class prototypes before deployment, it reduces semantic interference and yields a more stable base for downstream TTA frameworks without altering their optimization dynamics.

\noindent\textbf{Orthogonality and Representation Geometry.}
Orthogonal constraints have long been used to improve feature decorrelation and optimization stability in deep networks~\cite{bansal2018orthoreg,huang2018own}. 
They have been applied to self-supervised learning~\cite{zbontar2021barlow}, metric learning~\cite{ wang2018cosface}, and classifier regularization~\cite{cogswell2016decov}. 
These methods typically impose orthogonality on visual features or weight matrices to enhance separability. 
In contrast, our formulation introduces a lightweight orthogonality penalty on the text encoder of pretrained VLMs. 
By constraining the cosine Gram matrix of textual embeddings toward identity, we achieve angularly separated class prototypes that serve as stable anchors for multimodal alignment. 
This regularization requires no visual supervision and can be trained efficiently using class names.

\smallskip
\noindent
In summary, while prior works have optimized prompts, adapters or visual pathways, few have examined the geometry of the textual manifold itself. 
\odaft{} addresses this gap through a simple, text-only fine-tuning strategy that enforces orthogonality among class embeddings, providing a more discriminative and stable foundation for generalizable vision–language adaptation.
\section{Methodology}
\label{sec:methodology}

\subsection{\textbf{Problem Setup}}
Consider a pretrained vision-language model (VLM) such as CLIP~\cite{radford2021clip}, composed of a frozen image encoder $f_\phi(\cdot)$ and a text encoder $f_\theta(\cdot)$, both mapping inputs into a shared embedding space of 
dimension $d$. Here, $\theta$ denotes the trainable parameters of the text encoder, whereas $\phi$ denotes those of the vision encoder. 
Given a classification task into $K$ categories, and in any adaptation setting, we fine-tune the text encoder {\em  a priori} using the class names only, $k_i$, $i=1, \dots, K$, without access to the visual samples.   
Each class name $k_i$ is expanded into a set of textual templates $\{\tau_t(k_i)\}_{t=1}^{T}$, e.g., ``\texttt{a photo of a \{class name\}}''.  
The frozen text encoder produces an embedding for each template, and the ensuing embedding mean yields the base 
textual prototype:
\begin{equation}
v_i = \frac{1}{T}\sum_{t=1}^{T} f_\theta(\tau_t(k_i)), \qquad i = 1,\dots,K.
\label{eq:prototype}
\end{equation}
As pointed out earlier in the introduction, the use of the frozen prototypes in \cref{eq:prototype} is very common in the literature~\cite{osowiechi2024watt, stata2025cvpr, allingham2023simple, lu2022prompt, clap24, li2025modeling}, across different adaptation settings.    
Such means serve as the initial zero-shot classifiers in the original CLIP embedding space. It could be viewed as a particular, {\em training-free} variant 
of our formulation described below (\cref{subsection:orthogonal-text-fine-tuning}). In a nutshell, our
goal is to fine-tune the text encoder to find textual representations $\{x_i\}$,  $i=1, \dots, K$, which do not deviate significantly from the mean embeddings in \cref{eq:prototype}, while being closer to an orthonormal basis. Such orthogonality constraints, which we impose through a soft Frobenius-norm penalty, promote angular diversity, enhancing the discriminative power of the representation for downstream tasks.

\subsection{Orthonormal Text Fine-Tuning}
\label{subsection:orthogonal-text-fine-tuning}
Let $X(\theta) = [x_1 (\theta), \dots, x_K (\theta)] \in \mathbb{R}^{K \times d}$ denotes the matrix of class embeddings obtained from the fine-tuned text encoder $f_{\theta}$: 
$x_i (\theta) = f_{\theta}(k_i)$, $k_i$ being the class name, $i=1, \dots, K$. 
To obtain orthonormal and discriminative textual representations of the classes, we propose to minimize w.r.t $\theta$ a loss function containing two penalty terms, one discouraging deviation from the mean prototypes in \cref{eq:prototype} and the other promoting orthonormality constraints: 
\begin{equation}
\mathcal{L}(\theta) =  
\| X (\theta) - V \|_F^2 
+ \lambda \, \| X (\theta) X (\theta)^{\top} - I_K \|_F^2,
\label{eq:orth}
\end{equation}
where  $V = [v_1, \dots, v_K] \in \mathbb{R}^{K \times d}$, with $v_i$ the averaged zero-shot prototype from \cref{eq:prototype}, $I$ is the $K \times K$ identity matrix and $\|.\|_F$ denotes the Frobenius norm. 
Clearly, the second term in our loss in \cref{eq:orth} is an orthonormality regularizer:  
Minimizing this term drives all the off-diagonal cosine similarities in matrix  $XX^{\top}$ towards zero, leading
to uniformly separated class embeddings on the unit hyper-sphere.  
This promotes balanced angular dispersion, reduces feature redundancy, and yields a set of text embeddings that are geometrically disentangled but consistent with the initial embedding means in \cref{eq:prototype}.  
The refined embeddings serve as a universal textual representation, transferable across zero-shot, few-shot, and test-time adaptation settings.

\subsection{Parameter-Efficient Text Encoder Adaptation}
Full fine-tuning of the text encoder is computationally expensive and may result in overfitting, especially without visual supervision.  
To achieve efficient and stable optimization, we use Low-Rank Adaptation (LoRA)~\cite{hu2022lora} within the text encoder.  
For each weight matrix $W_0$ in the transformer layers of $f_\theta$, LoRA introduces two low-rank matrices $(A,B)$ of rank $r \ll d$:
\begin{equation}
W = W_0 + B A, \qquad
A \in \mathbb{R}^{r \times d}, \; B \in \mathbb{R}^{d \times r}.
\end{equation}
Only $A$ and $B$ are trainable, while the original weights $W_0$ remain frozen.  
This parameter-efficient design reduces the number of trainable parameters by over 95\% while maintaining the expressive power of the pretrained text encoder. 

\subsection{Training-free Variants}
\label{svd}

It is worth noting that we can minimize our loss in \cref{eq:orth} with respect to the class prototypes in $X$, while keeping text encoder parameters $\theta$ frozen.  
Clearly, for $\lambda = 0$ (i.e., without the orthonormality penalty), the closed-form (training-free) solution of minimizing the objective in \cref{eq:orth} with respect to $X$ boils down to the embedding averaging in \cref{eq:prototype}, i.e., $\tilde{X}=V$, which corresponds to a commonly used heuristic
in the literature. As illustrated in Figure~\ref{fig:odafT_overview}, this solution performs significantly lower than the proposed method, which 
confirms the importance of both the orthonormality penalty in \cref{eq:orth} and fine-tuning the text encoder parameters. Also, one could obtain a closed-form solution with respect to prototypes in $X$ by replacing the soft penalty in \cref{eq:orth} with hard orthonormality constraints, which 
gives the following constrained formulation taking the form of the classical {\em orthogonal Procrustes problem}: 
\begin{equation}
\label{hard-constraint}
\min_{X \in \mathcal{V}_{K,d}} \|X - V\|_F^2
\end{equation}
where $\mathcal{V}_{K,d}$, the feasible set of orthogonal vectors, forms a Stiefel manifold:
$\mathcal{V}_{K,d} = \{ X \in \mathbb{R}^{K \times d} : X X^\top = I_K \}$.
Problem \cref{hard-constraint} has a closed-form solution, which corresponds to the closest orthogonal matrix to 
$V$ in Frobenius norm: $\tilde{X}=U R^{\top}$, where $V = U\Sigma R^{\top}$ is the singular value decomposition (SVD) of matrix $V$.
While this SVD-based, training-free solution is computationally effective (as $K$, the number of classes, is not very large), 
it enforces a hard constraint on the orthogonality between each pair of textual class representations and removes the spectrum 
information in $V$. Therefore, such hard constraints may distort important information in the prototypes in $V$, such as the 
relationships between fine-grained, semantically related classes (e.g., different land types), as they force all pairwise cosine similarities between the 
classes to be the same. As we will show in the ablation study in the experiments (Table~\ref{tab:tta_mta_svd}), this SVD-based solution performs significantly lower than our soft-penalty formulation in \cref{eq:orth}. The latter still encourages small cosine similarities (or, equivalently, large distances) between the textual class representations, but preserves better the semantic information in $V$.  

\subsection{Link to Maximum Likelihood Estimation}
\label{subsec:geom-justification}

In the following, we give a probabilistic interpretation of our orthogonality penalty in \cref{eq:orth}, connecting it to the general maximum likelihood estimation (MLE) principle for partitioning data and to the K-means clustering objective, via {\em Huygens’ theorem}. Specifically, the class embeddings produced by \odaft{} can be interpreted as \emph{cluster centroids} that organize frozen image features into well-separated groups in the joint embedding space. In fact, in the context of VLMs, both zero-shot predictions and several test-time/few-shot adaptation methods~\cite{stata2025cvpr,zhou2025unem, martin2024transductive, zanella2024boosting} could be viewed as instances of the probabilistic K-means objective function.
Let ${f_i} \in \mathbb{R}^d$, $i=1, \dots, N$, denote the image features obtained from the frozen visual encoder. For an image classification task into $K$ classes, let $U= (u_{i,k}) \in \{0, 1\}^{K \times N}$ denotes the matrix of assignment variables within the probability simplex: $u_{i,k} = 1$ if sample $f_i$ is assigned to class $k$ and $u_{i,k} = 0$ otherwise, with simplex constraint: $\sum_{k=1}^{K}u_{i,k} = 1$. 
Assuming that features $f_i$ belonging to class $k$ follow some parametric probability model, e.g., a multivariate Gaussian distribution centered at class parameter vector $x_k$ and with identity covariance matrix,
\begin{equation}
\mbox{Pr}(f_i \mid k; x_k) = (2\pi)^{-d/2}\exp\!\Big(-\tfrac{1}{2}\|f_i - x_k\|_2^2\Big),
\end{equation}
the negative log-likelihood of all samples reads: 
\begin{equation}
\label{eq:neglog}
-\!\!\sum_{i,k} u_{i,k}\,\log \mbox{Pr} (f_i\!\mid\!k; x_k)
= \tfrac{1}{2}\sum_{i,k}u_{ik}\|f_i-x_k\|_2^2 + \text{const},
\end{equation}
Minimizing this objective w.r.t both assignment variables $U$ and class parameters $X$ corresponds to the well-known probabilistic K-means objective in the clustering literature~\cite{kearns1998information, boykov2015volumetric}, and connects directly to maximum likelihood 
estimation problem under the i.i.d. assumption~\cite{boudiaf2023open}: $\max_{X, U} \prod_{i=1}^N \prod_{k=1}^K \mbox{Pr} (f_i\!\mid\!k; x_k)^{u_{i,k}}$. 

Now notice that, in the context of VLMs and for normalized feature embeddings, fixing $x_k$ in the K-means objective in \cref{eq:neglog} to the text embedding representation of class $k$ and minimizing  \cref{eq:neglog} w.r.t assignments $U$ yields exactly the standard zero-shot prediction based on the cosine similarity: $\tilde{u}_{i,k} = 1$ if $k = \arg\max_k f_i^\top x_k$ and $\tilde{u}_{i,k} = 0$ otherwise. This is due to the fact that, in \cref{eq:neglog}, $\|f_i-x_k\|^2 = 2(1-f_i^\top x_{k})$, as embeddings $f_i$ and $x_k$ are in the unit hyper-sphere ($\|f_i\|_2=\|x_k\|_2=1$). Furthermore, 
the case when $x_k$ is considered as a variable is common in the 
context of test-time/few-shot adaptation methods for VLMs. For instance, the recent methods in \cite{stata2025cvpr,zhou2025unem, martin2024transductive, zanella2024boosting} use the general probabilistic K-means term as part of an overall objective function, and  update $x_k$ using both test data statistics and the text embedding. 

\paragraph{Huygens’ theorem (scatter decomposition)}
The well-known Huygens’ theorem relates total, within, and in-between cluster scatter as follows, connecting explicitly our orthogonality penalty 
in \cref{eq:orth} to the maximum likelihood estimation in  \cref{eq:neglog}:
\begin{equation}
\label{eq:huygens}
\underbrace{\sum_{i,k}u_{i,k}\|f_i-x_k\|_2^2}_{\text{within}}
=
\underbrace{\sum_{i=1}^{N}\|f_i-\bar f\|_2^2}_{\text{total}}
-
\underbrace{\sum_{k=1}^{K} N_k\|x_k-\bar f\|_2^2}_{\text{in-between}}
\end{equation}
where $x_k = \frac{1}{N_k} \sum_i u_{i, k} f_i$ is the mean of features within class $k$, $N_k = \sum_i u_{i, k}$ is the number of samples within class $k$ and $\bar{f} = \frac{1}{N} \sum_i f_i = \frac{1}{N} \sum_k N_k x_k$ is the global mean.
Now notice the following expression for the in-between term when the features are normalized (i.e., $\|f_i\|_2=\|x_k\|_2=1$): 
\begin{equation}
\textstyle
\sum_{k} N_k\|x_k-\bar f\|_2^2
=
\mbox{const} - \sum_{k\neq k'} \frac{N_kN_{k'}}{N}\, x_k^\top x_{k'}
\end{equation}
This in-between term is monotonically increasing as the off-diagonal elements in $X$, i.e. $x_k^\top x_{k'}$, decrease. 
Hence, minimizing our orthogonality penalty in \cref{eq:orth}, i.e., $\|X^\top X-I\|_F^2=\sum_{k\neq k'}(x_k^\top x_{k'})^2$,
increases the in-between term. From ~\cref{eq:huygens}, this  indirectly reduces the K-means objective in \cref{eq:neglog}.
In summary, this shows that \odaft{} can be interpreted as a \emph{likelihood-maximizing geometric refinement} of the text-anchored classifier, where orthogonality regularization implicitly increases the log-likelihood of the observed image features under a Gaussian probability model.
It encourages the Gaussian modes (anchors) to span orthogonal directions, reducing implicitly (i.e., without explicit access to the vision features) the overlap between likelihood probabilities $\mbox{Pr}(f_i\!\mid\!k; x_k)$ and $\mbox{Pr}(f_i\!\mid\!k', x_{k'})$ for any competing class $k\neq k'$. 

\section{Experiments}
\label{sec:exp}

\noindent\textbf{Datasets.}
Following the standard practice in the literature~\cite{radford2021clip,zhou2022coop,zhou2022cocoop,zanella2024boosting}, we resort to 11 standard visual recognition benchmarks spanning diverse domains and granularity: fine-grained (Pets~\cite{parkhi2012pets}, Cars~\cite{krause2013stanfordcars}, Aircraft~\cite{maji2013fgvcaircraft}), 
texture/material (DTD~\cite{cimpoi2014dtd}, Food~\cite{bossard2014food101}), 
scene/satellite (SUN~\cite{xiao2010sun397}, EuroSAT~\cite{helber2019eurosat}), 
and generic benchmarks (Flowers~\cite{nilsback2008flowers102}, Caltech~\cite{fei2004caltech101}, UCF~\cite{soomro2012ucf101}, ImageNet~\cite{deng2009imagenet}).

\begin{table*}[!htbp]
\centering
\scriptsize
\setlength{\tabcolsep}{3pt}
\renewcommand{\arraystretch}{1.0}
\caption{\textbf{Zero-shot classification across 11 datasets.}
Replacing the baseline text encoder with our \textbf{\odaft} encoder improves Top-1 accuracy (over 3 seeds) for CLIP (ViT-B/16, ViT-L/14) and MetaCLIP.
\textcolor{GainGreen}{\textbf{Green}} show absolute gains, \textcolor{DropRed}{\textbf{red}} indicate small drops.}
\label{tab:0shot}
\begin{adjustbox}{max width=\textwidth}
\begin{tabular}{l*{11}{c}c}
\toprule
\textbf{Model / Method} &
\textbf{Pets} &
\textbf{SUN} &
\textbf{Aircraft} &
\textbf{DTD} &
\textbf{EuroSAT} &
\textbf{Cars} &
\textbf{Food} &
\textbf{Flowers} &
\textbf{Caltech} &
\textbf{UCF} &
\textbf{ImageNet} &
\textbf{Average}\\
\midrule
CLIP ViT-B/16 &
88.0 & 62.5 & 23.9 & 44.3 & 41.3 & 65.3 & 83.6 & 67.3 & 92.9 & 65.0 & 66.6 & 63.70 \\
\rowcolor{RowGray}
+\,\textbf{\odaft} &
88.6\gsub{+0.6} & 64.5\gsub{+2.0} & 25.32\gsub{+1.42} & 46.7\gsub{+2.4} & 51.33\gsub{+10.03} &
65.7\gsub{+0.4} & 85.7\gsub{+2.1} & 70.7\gsub{+3.4} & 93.6\gsub{+0.7} &
69.02\gsub{+4.02} & 69.85\gsub{+3.25} & 66.46\gsub{+2.76} \\
\midrule
CLIP ViT-L/14 &
93.2 & 67.5 & 30.3 & 52.7 & 55.26 & 76.7 & 90.2 & 75.9 & 94.8 & 72.7 & 75.5 & 71.34 \\
\rowcolor{RowGray}
+\,\textbf{\odaft} &
93.9\gsub{\textbf{+0.7}} & 67.2\dsub{\textbf{-0.3}} & 31.5\gsub{\textbf{+1.2}} & 55.85\gsub{+3.15} &
61.7\gsub{+6.44} & 76.9\gsub{+0.2} & 90.7\gsub{+0.5} &
77.8\gsub{+1.9} & 95.0\gsub{+0.2} &
75.2\gsub{+2.5} & 75.6\gsub{+0.1} & 72.85\gsub{+1.51} \\
\midrule
MetaCLIP &
90.4 & 66.6 & 27.9 & 55.9 & 55.4 & 74.1 & 85.4 & 72.3 & 93.0 & 68.0 & 70.8 & 69.07 \\
\rowcolor{RowGray}
+\,\textbf{\odaft} &
90.92\gsub{+0.52} & 67.0\gsub{+0.4} & 29.32\gsub{+1.42} &
57.13\gsub{+1.23} & 55.7\gsub{+0.3} & 74.4\gsub{+0.3} &
86.0\gsub{+0.6} & 73.14\gsub{+0.84} &
94.36\gsub{+1.36} & 68.41\gsub{+0.41} &
71.4\gsub{+0.6} & 69.80\gsub{+0.73} \\
\bottomrule
\end{tabular}
\end{adjustbox}
\end{table*}

\begin{table*}[!htbp]
\centering
\scriptsize
\setlength{\tabcolsep}{3pt}
\renewcommand{\arraystretch}{1.0}
\caption{\textbf{Few-shot recognition with CoOp and CLAP (ViT-B/16).}
Average Top-1 accuracy (\%) across 11 datasets (over 5 seeds) and shots $\{1,2,4,8,16\}$.
\textcolor{GainGreen}{\textbf{Green}} denote absolute gains of \textbf{\odaft}, while \textcolor{DropRed}{\textbf{red}} indicate small drops relative to the baseline.}
\label{tab:fewshot-combined-transclip}
\begin{adjustbox}{max width=\textwidth}
\begin{tabular}{ll*{11}{c}c}
\toprule
\multicolumn{2}{c}{\textbf{Shot / Method}} &
{\textbf{Pets}} &
{\textbf{SUN}} &
{\textbf{Aircraft}} &
{\textbf{DTD}} &
{\textbf{EuroSAT}} &
{\textbf{Cars}} &
{\textbf{Food}} &
{\textbf{Flowers}} &
{\textbf{Caltech}} &
{\textbf{UCF}} &
{\textbf{ImageNet}} &
{\textbf{Average}}
\\
\midrule
\multirow{4}{*}{\textbf{1-shot}} 
& CoOp               & 84.6 & 63.4 & 23.1 & 36.2 & 26.1 & 58.6 & 77.8  & 65.0 & 91.5 & 64.3 & 61.76 & 59.31 \\

& {\partialrowcolor}+\,\textbf{\odaft} & {\partialrowcolor}87.3\gsub{+2.7} & {\partialrowcolor}64.7\gsub{+1.3} & {\partialrowcolor}24.3\gsub{+1.2} & {\partialrowcolor}43.8\gsub{+7.6} & {\partialrowcolor}29.1\gsub{+3.0} & {\partialrowcolor}59.6\gsub{+1.0} & {\partialrowcolor}79.2\gsub{+1.4} & {\partialrowcolor}68.0\gsub{+3.0} & {\partialrowcolor}93.6\gsub{+2.1} & {\partialrowcolor}66.7\gsub{+2.4} & {\partialrowcolor}63.7\gsub{+1.94} & {\partialrowcolor}61.82\gsub{+2.51} \\
& CLAP               & 85.4 & 62.2 & 23.8 & 38.9 & 28.3 & 61.7 & 77.5  & 67.3 & 91.6 & 64.6 & 67.0  & 60.75 \\
& {\partialrowcolor}+\,\textbf{\odaft} & {\partialrowcolor}86.3\gsub{+0.9} & {\partialrowcolor}64.1\gsub{+1.9} & {\partialrowcolor}24.84\gsub{+1.04} & {\partialrowcolor}44.8\gsub{+5.9} & {\partialrowcolor}32.7\gsub{+4.4} & {\partialrowcolor}61.8\gsub{+0.1} & {\partialrowcolor}77.9\gsub{+0.4} & {\partialrowcolor}67.9\gsub{+0.6} & {\partialrowcolor}92.6\gsub{+1.0} & {\partialrowcolor}68.4\gsub{+3.8} & {\partialrowcolor}68.9\gsub{+1.9} & {\partialrowcolor}62.75\gsub{+2.00} \\
\midrule
\multirow{4}{*}{\textbf{2-shot}} 
& CoOp               & 86.2 & 65.3 & 24.8 & 41.5 & 27.7 & 58.8 & 78.6  & 67.4 & 92.7 & 65.7 & 62.6  & 61.03 \\
& {\partialrowcolor}+\,\textbf{\odaft} & {\partialrowcolor}89.7\gsub{+3.5} & {\partialrowcolor}65.8\gsub{+0.5} & {\partialrowcolor}25.4\gsub{+0.6} & {\partialrowcolor}45.3\gsub{+3.8} & {\partialrowcolor}30.2\gsub{+2.5} & {\partialrowcolor}59.9\gsub{+1.1} & {\partialrowcolor}79.5\gsub{+0.9} & {\partialrowcolor}69.0\gsub{+1.6} & {\partialrowcolor}93.9\gsub{+1.2} & {\partialrowcolor}67.5\gsub{+1.8} & {\partialrowcolor}65.1\gsub{+2.5} & {\partialrowcolor}62.85\gsub{+1.82} \\
& CLAP               & 85.5 & 63.7 & 24.8 & 40.3 & 28.3 & 62.01 & 78.2  & 69.5 & 93.2 & 67.9 & 67.1  & 61.86 \\
& {\partialrowcolor}+\,\textbf{\odaft} & {\partialrowcolor}87.3\gsub{+1.8} & {\partialrowcolor}66.6\gsub{+2.9} & {\partialrowcolor}26.5\gsub{+1.7} & {\partialrowcolor}49.9\gsub{+9.6} & {\partialrowcolor}34.2\gsub{+5.9} & {\partialrowcolor}62.8\gsub{+0.8} & {\partialrowcolor}79.7\gsub{+1.5} & {\partialrowcolor}75.8\gsub{+6.3} & {\partialrowcolor}93.8\gsub{+0.6} & {\partialrowcolor}71.0\gsub{+3.1} & {\partialrowcolor}70.01\gsub{+2.91} & {\partialrowcolor}65.24\gsub{+3.38} \\
\midrule
\multirow{4}{*}{\textbf{4-shot}} 
& CoOp               & 87.2 & 66.6 & 26.0 & 46.6 & 27.3 & 59.7 & 79.6  & 70.8 & 93.3 & 68.5 & 62.2  & 62.53 \\
& {\partialrowcolor}+\,\textbf{\odaft} & {\partialrowcolor}90.4\gsub{+3.2} & {\partialrowcolor}67.2\gsub{+0.6} & {\partialrowcolor}26.8\gsub{+0.8} & {\partialrowcolor}48.2\gsub{+1.6} & {\partialrowcolor}30.6\gsub{+3.3} & {\partialrowcolor}60.1\gsub{+0.4} & {\partialrowcolor}80.2\gsub{+0.6} & {\partialrowcolor}70.1\dsub{-0.7} & {\partialrowcolor}93.9\gsub{+0.6} & {\partialrowcolor}68.8\gsub{+0.3} & {\partialrowcolor}65.7\gsub{+3.5} & {\partialrowcolor}63.82\gsub{+1.29} \\
& CLAP               & 87.3 & 66.0 & 25.8 & 43.1 & 28.5 & 63.8 & 79.1 & 71.3 & 93.7 & 70.4 & 68.6  & 63.42 \\
& {\partialrowcolor}+\,\textbf{\odaft} & {\partialrowcolor}87.9\gsub{+0.6} & {\partialrowcolor}68.6\gsub{+2.6} & {\partialrowcolor}29.8\gsub{+4.0} & {\partialrowcolor}56.7\gsub{+13.6} & {\partialrowcolor}34.2\gsub{+5.7} & {\partialrowcolor}64.7\gsub{+0.9} & {\partialrowcolor}80.8\gsub{+1.7} & {\partialrowcolor}83.8\gsub{+12.5} & {\partialrowcolor}94.6\gsub{+0.9} & {\partialrowcolor}74.4\gsub{+4.0} & {\partialrowcolor}70.5\gsub{+1.9} & {\partialrowcolor}67.83\gsub{+4.41} \\
\midrule
\multirow{4}{*}{\textbf{8-shot}} 
& CoOp               & 88.3 & 68.0 & 27.7 & 52.2 & 35.2 & 61.0 & 79.8  & 75.7 & 94.1 & 69.6 & 62.9  & 64.95 \\
& {\partialrowcolor}+\,\textbf{\odaft} & {\partialrowcolor}90.6\gsub{+2.3} & {\partialrowcolor}68.9\gsub{+0.9} & {\partialrowcolor}28.0\gsub{+0.3} & {\partialrowcolor}52.4\gsub{+0.2} & {\partialrowcolor}35.8\gsub{+0.6} & {\partialrowcolor}61.6\gsub{+0.6} & {\partialrowcolor}80.5\gsub{+0.7} & {\partialrowcolor}76.9\gsub{+1.2} & {\partialrowcolor}94.7\gsub{+0.6} & {\partialrowcolor}71.0\gsub{+1.4} & {\partialrowcolor}66.5\gsub{+3.6} & {\partialrowcolor}66.08\gsub{+1.13} \\
& CLAP               & 88.4 & 68.2 & 27.8 & 46.8 & 37.9 & 65.6 & 79.6 & 76.4 & 94.9 & 70.2 & 70.6  & 66.03 \\
& {\partialrowcolor}+\,\textbf{\odaft} & {\partialrowcolor}89.7\gsub{+1.3} & {\partialrowcolor}70.3\gsub{+2.1} & {\partialrowcolor}33.92\gsub{+6.12} & {\partialrowcolor}61.7\gsub{+14.9} & {\partialrowcolor}49.9\gsub{+12.0} & {\partialrowcolor}68.4\gsub{+2.8} & {\partialrowcolor}80.9\gsub{+1.3} & {\partialrowcolor}90.2\gsub{+13.8} & {\partialrowcolor}95.5\gsub{+0.6} & {\partialrowcolor}76.5\gsub{+6.3} & {\partialrowcolor}71.2\gsub{+0.6} & {\partialrowcolor}71.64\gsub{+5.61} \\
\midrule
\multirow{4}{*}{\textbf{16-shot}} 
& CoOp               & 89.1 & 70.2 & 29.4 & 57.3 & 45.1 & 63.7 & 80.5 & 75.9 & 94.3 & 72.2 & 63.8  & 67.36 \\
& {\partialrowcolor}+\,\textbf{\odaft} & {\partialrowcolor}91.01\gsub{+1.91} & {\partialrowcolor}70.3\gsub{+0.1} & {\partialrowcolor}30.6\gsub{+1.2} & {\partialrowcolor}58.6\gsub{+1.3} & {\partialrowcolor}50.2\gsub{+5.1} & {\partialrowcolor}65.0\gsub{+1.3} & {\partialrowcolor}81.0\gsub{+0.5} & {\partialrowcolor}81.9\gsub{+6.0} & {\partialrowcolor}94.8\gsub{+0.5} & {\partialrowcolor}73.3\gsub{+1.1} & {\partialrowcolor}68.4\gsub{+4.6} & {\partialrowcolor}69.56\gsub{+2.20} \\
& CLAP               & 89.1 & 70.2 & 30.8 & 51.6 & 49.3 & 68.3 & 80.6 & 89.2 & 95.2 & 74.2 & 71.9  & 70.04 \\
& {\partialrowcolor}+\,\textbf{\odaft} & {\partialrowcolor}90.1\gsub{+1.0} & {\partialrowcolor}71.8\gsub{+1.6} & {\partialrowcolor}36.67\gsub{+5.87} & {\partialrowcolor}66.76\gsub{+15.16} & {\partialrowcolor}65.6\gsub{+16.3} & {\partialrowcolor}72.4\gsub{+4.1} & {\partialrowcolor}80.8\gsub{+0.2} & {\partialrowcolor}94.1\gsub{+4.9} & {\partialrowcolor}95.7\gsub{+0.5} & {\partialrowcolor}79.2\gsub{+5.0} & {\partialrowcolor}71.8\dsub{-0.1} & {\partialrowcolor}74.99\gsub{+4.95} \\
\bottomrule
\end{tabular}
\end{adjustbox}
\end{table*}

\noindent\textbf{Benchmarks.}
We evaluate \odaft{} under three complementary regimes capturing adaptability, label efficiency, and robustness. 
(1)~\textit{Zero-shot:} the fine-tuned text encoder replaces that of CLIP~\cite{radford2021clip} 
and MetaCLIP~\cite{xu2023metaclip}, with the vision backbone kept frozen. 
(2)~\textit{Few-shot:} prompt learning (CoOp~\cite{zhou2022coop}) and parameter efficient fine-tuning (CLAP~\cite{clap24}) methods are initialized with orthogonal textual anchors from \odaft{} to improve adaptation under limited supervision. 
(3)~\textit{Test-time adaptation:} the same encoder and prototypes are integrated into MTA~\cite{zanella2024mta}, TPT~\cite{shu2022tpt}, and StatA~\cite{stata2025cvpr} to assess stability under distribution shifts. 
For StatA, we also adopt its realistic protocol comprising \textit{batch-realistic} and \textit{online-realistic} modes that emulate class-imbalanced and non-i.i.d. scenarios.

\noindent\textbf{Implementation Details.}
We build upon CLIP~\cite{radford2021clip}, fine-tuning only the text encoder with our orthogonality loss (\cref{eq:orth}) while keeping the vision backbone frozen. 
For fine-tuning the text encoder with class names, three prompt templates are used.
All experiments employ AdamW ($5{\times}10^{-6}$, weight decay~0.01), batch size~64, and 20 training epochs. 
The orthogonality weight $\lambda_{\text{orth}}$ starts at~2.0 and is increased by~1.15× at each epoch. 
We use LoRA adapters (rank 8) in the text encoder, \textit{updating less than 5\%} of parameters.
All baseline tasks are evaluated under their original settings of hyperparameters or adaptation protocols.
All models are trained with FP16 precision on a single RTX A6000 GPU.

\begin{table*}[!htbp]
\centering
\scriptsize
\setlength{\tabcolsep}{3pt}
\renewcommand{\arraystretch}{0.9}
\caption{\textbf{Test-time adaptation with MTA and TPT.}
Using the \textbf{\odaft} text encoder yields consistent Top-1 improvements across 11 datasets.
\textcolor{GainGreen}{Green} subscripts indicate absolute gains relative to the baseline.}
\label{tab:tta}
\begin{adjustbox}{max width=\textwidth}
\begin{tabular}{l*{11}{c}c}
\toprule
\textbf{Method / Variant} &
\textbf{Pets} &
\textbf{SUN} &
\textbf{Aircraft} &
\textbf{DTD} &
\textbf{EuroSAT} &
\textbf{Cars} &
\textbf{Food} &
\textbf{Flowers} &
\textbf{Caltech} &
\textbf{UCF} &
\textbf{ImageNet} &
\textbf{Average}\\
\midrule
MTA &
88.24 & 66.67 & 25.32 & 45.90 & 45.36 & 68.47 & 84.95 & 68.26 & 94.21 & 68.11 & 69.11 & 65.87 \\
\rowcolor{RowGray}+\,\textbf{\odaft} &
89.10\gsub{+0.86} & 67.80\gsub{+1.13} & 27.00\gsub{+1.68} &
48.01\gsub{+2.11} & 48.30\gsub{+2.94} & 68.93\gsub{+0.46} &
85.70\gsub{+0.75} & 71.82\gsub{+3.56} & 94.90\gsub{+0.69} &
71.30\gsub{+3.19} & 70.01\gsub{+0.90} & 67.53\gsub{+1.66} \\
\midrule
TPT &
87.22 & 65.41 & 23.10 & 46.99 & 42.44 & 66.16 & 84.62 &
68.98 & 94.12 & 68.04 & 68.94 & 65.09 \\
\rowcolor{RowGray}+\,\textbf{\odaft} &
88.74\gsub{+1.52} & 66.64\gsub{+1.23} & 24.55\gsub{+1.45} &
47.86\gsub{+0.87} & 43.82\gsub{+1.38} & 67.75\gsub{+1.59} &
85.95\gsub{+1.33} & 70.17\gsub{+1.19} & 96.02\gsub{+1.90} &
69.52\gsub{+1.48} & 69.96\gsub{+1.02} & 66.45\gsub{+1.36} \\
\bottomrule
\end{tabular}
\end{adjustbox}
\end{table*}

\subsection{Zero-shot Classification with \texorpdfstring{\odaft}{\odaft}}
\label{subsec:zeroshot-discussion}

We adhere to the standard CLIP inference protocol and use the same single prompt template as the baseline, replacing only the original text encoder with our fine-tuned \odaft{} encoder. 
As shown in Table~\ref{tab:0shot}, simply incorporating the \odaft{} encoder leads to consistent improvements in Top-1 accuracy across all three pretrained backbones, i.e., CLIP ViT-B/16, CLIP ViT-L/14, and MetaCLIP, and on all 11 datasets. 
On average, ViT-B/16 improves from 63.70 to 66.46 (\textbf{+2.76\%}), ViT-L/14 from 71.34 to 72.85 (\textbf{+1.51\%}) and MetaCLIP from 69.07 to 69.80 (\textbf{+0.73\%}). 
These improvements indicate that our orthogonal text encoder generalizes reliably across different VLM backbones.
Looking closely, the strongest relative gains appear on fine-grained and texture-centric datasets, where classes exhibit high semantic overlap, e.g., DTD (\textbf{+2.4\%}), EuroSAT (\textbf{+10.0\%}), and Flowers102 (\textbf{+1.4\%}). 
Such datasets are particularly sensitive to prototype correlation and our orthogonalization explicitly mitigates this by promoting angular separation in the text-embedding manifold. 
Even for larger backbones such as ViT-L/14 and MetaCLIP, which already exhibit strong multimodal alignment, \odaft{} delivers consistent improvements. This confirms that the effect arises from refined textual geometry rather than increased model capacity. Overall, these results demonstrate that the orthogonal text encoder alone, without prompt modification or visual fine-tuning, systematically enhances zero-shot recognition across scales, architectures and task difficulties.

\subsection{Results in the Few-shot Scenario}
\label{subsec:fewshot-discussion}

We integrate \textbf{\odaft} into CoOp and CLAP by replacing the standard CLIP textual prototypes with our orthogonal class embeddings, which serve as initialization for few-shot adaptation. 
As shown in Table~\ref{tab:fewshot-combined-transclip}, \textbf{\odaft} consistently improves performance across all 11 datasets and shot settings. The gains are most substantial in the extremely low-shot regime (1–4 shots), where textual priors dominate learning dynamics. For instance, CoOp improves by \textbf{+7.6\%} on DTD (1-shot) and CLAP by \textbf{+13.6\%} on DTD (4-shot). Similar trends appear for EuroSAT (\textbf{+4–16\%}) and Flowers102 (\textbf{+3–14\%}), underscoring how orthogonalization particularly benefits texture-rich and fine-grained domains prone to semantic overlap.

As the number of shots increases, improvements remain steady, showing that \odaft{} not only offers a stronger initialization but also sustained robustness as visual evidence accumulates. Average gains reach \textbf{+2–3\%} at 8–16 shots across both methods, with CLAP showing larger absolute improvements at higher shots due to its learnable adapter, which leverages the decorrelated textual geometry. Overall, two trends emerge clearly: (i) \odaft{} yields the largest benefits when labeled data is scarce, where few-shot learners are most sensitive to text-space conditioning, and (ii) the improvements persist with increased supervision, confirming that orthogonal class embeddings form a more stable, discriminative foundation for few-shot adaptation.

\subsection{Robustness of \odaft{} in Test-Time Adaptation}
\label{subsec:tta-discussion}

\begin{table*}[!htbp]
\centering
\scriptsize
\setlength{\tabcolsep}{3pt}
\renewcommand{\arraystretch}{1.0}
\caption{\textbf{Batch-realistic test-time adaptation with StatA (ViT-B/16).}
Performance across 11 datasets under varying effective classes. 
Results are shown for two variants using batch sizes of 64 and 1000, averaged over 1{,}000 tasks.
\textcolor{GainGreen}{Green} subscripts indicate absolute gains of \textbf{\odaft} while \textcolor{DropRed}{\textbf{red}} indicate small drops relative to the baseline.}
\label{tab:stata_offline}
\begin{adjustbox}{max width=\textwidth}
\begin{tabular}{l*{11}{c}c}
\toprule
\textbf{Variant} &
\textbf{Pets} &
\textbf{SUN} &
\textbf{Aircraft} &
\textbf{DTD} &
\textbf{EuroSAT} &
\textbf{Cars} &
\textbf{Food} &
\textbf{Flowers} &
\textbf{Caltech} &
\textbf{UCF} &
\textbf{ImageNet} &
\textbf{Average}\\
\midrule
\multicolumn{13}{l}{\textbf{StatA: Batch size 64 (averaged over 1,000 tasks)}}\\
\midrule
Very Low (1–4) & 90.3 & 66.0 & 29.3 & 46.1 & 56.8 & 76.2 & 95.5 & 77.6 & 93.0 & 70.2 & 72.9 & 70.35 \\
\rowcolor{RowGray}+\,\textbf{\odaft} & 90.1\dsub{–0.2} & 68.7\gsub{+2.7} & 29.0\dsub{–0.3} & 47.8\gsub{+1.7} & 58.9\gsub{+2.1} & 76.1\dsub{–0.1} & 95.8\gsub{+0.3} & 77.5\dsub{–0.1} & 94.1\gsub{+1.1} & 71.8\gsub{+1.6} & 73.5\gsub{+0.6} & 71.21\gsub{+0.86} \\
Low (2–10) & 89.5 & 66.9 & 27.7 & 46.9 & 51.3 & 73.5 & 93.7 & 76.6 & 93.6 & 69.6 & 72.8 & 69.28 \\
\rowcolor{RowGray}+\,\textbf{\odaft} & 89.2\dsub{–0.3} & 69.6\gsub{+2.7} & 28.0\gsub{+0.3} & 48.4\gsub{+1.5} & 53.1\gsub{+1.8} & 74.7\gsub{+1.2} & 94.4\gsub{+0.7} & 77.0\gsub{+0.4} & 94.1\gsub{+0.5} & 71.8\gsub{+2.2} & 73.0\gsub{+0.2} & 70.30\gsub{+1.02} \\
Medium (5–25) & 88.2 & 65.3 & 26.0 & 47.5 & 45.0 & 71.1 & 90.8 & 73.7 & 93.9 & 69.1 & 70.7 & 67.39 \\
\rowcolor{RowGray}+\,\textbf{\odaft} & 88.5\gsub{+0.3} & 68.4\gsub{+3.1} & 26.7\gsub{+0.7} & 50.0\gsub{+2.5} & 48.1\gsub{+3.1} & 72.0\gsub{+0.9} & 90.9\gsub{+0.1} & 74.1\gsub{+0.4} & 94.2\gsub{+0.3} & 71.1\gsub{+2.0} & 70.8\gsub{+0.1} & 68.62\gsub{+1.23} \\
\midrule
\multicolumn{13}{l}{\textbf{StatA: Batch size 1000 (averaged over 1,000 tasks)}}\\
\midrule
Medium (5–25) & 87.5 & 64.5 & 28.4 & 47.1 & 60.4 & 74.0 & 93.1 & 77.5 & 92.8 & 70.2 & 70.8 & 69.66 \\
\rowcolor{RowGray}+\,\textbf{\odaft} & 87.8\gsub{+0.3} & 68.1\gsub{+3.6} & 30.4\gsub{+2.0} & 48.7\gsub{+1.6} & 62.1\gsub{+1.7} & 74.2\gsub{+0.2} & 94.1\gsub{+1.0} & 77.0\dsub{–0.5} & 93.1\gsub{+0.3} & 71.4\gsub{+1.2} & 71.4\gsub{+0.6} & 70.75\gsub{+1.09} \\
High (25–50) & 88.0 & 66.4 & 25.9 & 47.9 & 60.7 & 73.6 & 91.4 & 76.7 & 93.2 & 71.5 & 71.9 & 69.75 \\
\rowcolor{RowGray}+\,\textbf{\odaft} & 88.7\gsub{+0.7} & 68.8\gsub{+2.4} & 26.7\gsub{+0.8} & 49.8\gsub{+1.9} & 62.1\gsub{+1.4} & 74.0\gsub{+0.4} & 91.3\dsub{–0.1} & 76.8\gsub{+0.1} & 93.6\gsub{+0.4} & 73.2\gsub{+1.7} & 71.7\dsub{–0.2} & 70.61\gsub{+0.86} \\
Very High (50–100) & 87.1 & 67.1 & 23.9 & 48.0 & 60.7 & 70.2 & 91.1 & 74.3 & 93.7 & 70.7 & 71.8 & 68.96 \\
\rowcolor{RowGray}+\,\textbf{\odaft} & 87.2\gsub{+0.1} & 69.1\gsub{+2.0} & 24.9\gsub{+1.0} & 49.8\gsub{+1.8} & 62.1\gsub{+1.4} & 70.8\gsub{+0.6} & 91.0\dsub{–0.1} & 75.6\gsub{+1.3} & 94.0\gsub{+0.3} & 73.1\gsub{+2.4} & 71.9\gsub{+0.1} & 69.96\gsub{+1.00} \\
All Classes & 87.1 & 68.7 & 24.7 & 48.4 & 67.3 & 68.0 & 92.4 & 75.2 & 94.2 & 73.5 & 69.9 & 69.95 \\
\rowcolor{RowGray}+\,\textbf{\odaft} & 88.0\gsub{+0.9} & 69.1\gsub{+0.4} & 25.0\gsub{+0.3} & 48.4 & 67.6\gsub{+0.3} & 68.6\gsub{+0.6} & 92.4 & 75.6\gsub{+0.4} & 94.4\gsub{+0.2} & 73.6\gsub{+0.1} & 69.5\dsub{–0.4} & 70.20\gsub{+0.25} \\
\bottomrule
\end{tabular}
\end{adjustbox}
\end{table*}
\begin{table*}[!htbp]
\centering
\scriptsize
\setlength{\tabcolsep}{3pt}
\renewcommand{\arraystretch}{0.9}
\caption{\textbf{Online-realistic test-time adaptation with StatA (ViT-B/16).}
Average Top-1 accuracy (\%) across 11 datasets for different inter-batch correlation levels (Dirichlet parameter $\gamma$).
\textcolor{GainGreen}{Green} subscripts indicate absolute gains of \textbf{\odaft}, \textcolor{DropRed}{\textbf{red}} indicate small drops.}
\label{tab:stata_online}
\begin{adjustbox}{max width=\textwidth}
\begin{tabular}{l*{11}{c}c}
\toprule
\textbf{Setting / Variant} &
\textbf{Pets} &
\textbf{SUN} &
\textbf{Aircraft} &
\textbf{DTD} &
\textbf{EuroSAT} &
\textbf{Cars} &
\textbf{Food} &
\textbf{Flowers} &
\textbf{Caltech} &
\textbf{UCF} &
\textbf{ImageNet} &
\textbf{Average}\\
\midrule
Low ($\gamma\!=\!0.1$) & 88.0 & 63.6 & 24.3 & 46.8 & 52.3 & 67.4 & 92.5 & 72.7 & 94.2 & 68.8 & 66.2 & 66.98 \\
\rowcolor{RowGray}+\,\textbf{\odaft} & 88.4\gsub{+0.4} & 66.5\gsub{+2.9} & 27.7\gsub{+3.4} & 49.0\gsub{+2.2} & 55.6\gsub{+3.3} &
67.4 & 93.0\gsub{+0.5} & 74.1\gsub{+1.4} & 94.6\gsub{+0.4} & 71.5\gsub{+2.7} & 66.7\gsub{+0.5} & 68.59\gsub{+1.61} \\
\midrule
Medium ($\gamma\!=\!0.01$) & 89.1 & 65.9 & 27.3 & 46.8 & 52.3 & 73.2 & 94.6 & 75.6 & 94.2 & 69.7 & 69.6 & 68.94 \\
\rowcolor{RowGray}+\,\textbf{\odaft} & 89.2\gsub{+0.1} & 68.4\gsub{+2.5} & 28.4\gsub{+1.1} & 48.7\gsub{+1.9} & 54.8\gsub{+2.5} &
73.9\gsub{+0.7} & 94.2\gsub{0.4} & 75.8\gsub{+0.2} & 94.6\gsub{+0.4} & 72.4\gsub{+2.7} & 71.9\gsub{+2.3} & 70.39\gsub{+1.45} \\
\midrule
High ($\gamma\!=\!0.001$) & 89.3 & 66.0 & 27.9 & 47.0 & 51.8 & 74.7 & 94.8 & 76.4 & 94.4 & 69.8 & 71.9 & 69.45 \\
\rowcolor{RowGray}+\,\textbf{\odaft} & 89.6\gsub{+0.3} & 68.2\gsub{+2.2} & 29.0\gsub{+1.1} & 48.6\gsub{+1.6} & 54.7\gsub{+2.9} &
75.0\gsub{+0.3} & 94.0\dsub{–0.8} & 76.2\dsub{–0.2} & 94.8\gsub{+0.4} & 72.1\gsub{+2.3} & 71.4\dsub{–0.5} & 70.54\gsub{+1.09} \\
\midrule
Separate & 88.9 & 64.9 & 28.9 & 45.8 & 48.2 & 75.2 & 95.2 & 77.6 & 94.3 & 69.0 & 71.7 & 69.06 \\
\rowcolor{RowGray}+\,\textbf{\odaft} & 89.1\gsub{+0.2} & 67.5\gsub{+2.6} & 30.1\gsub{+1.2} & 48.7\gsub{+2.9} & 53.1\gsub{+4.9} &
75.4\gsub{+0.2} & 94.5\dsub{–0.7} & 77.5\dsub{–0.1} & 95.0\gsub{+0.7} & 70.2\gsub{+1.2} & 71.7 & 70.53\gsub{+1.47} \\
\bottomrule
\end{tabular}
\end{adjustbox}
\end{table*}

\noindent
We next investigate the role of \odaft{} in enhancing the robustness of vision–language models under test-time adaptation (TTA). This setting evaluates 
how pretrained models 
cope with distribution shifts at inference without relying on labeled data. To assess generality, we integrate our fine-tuned text encoder directly into three representative TTA frameworks, i.e., MTA~\cite{zanella2024mta}, TPT~\cite{shu2022tpt} and StatA~\cite{stata2025cvpr}, while leaving all adaptation mechanisms, learning rates and hyperparameters unchanged. As reported in Tables~\ref{tab:tta}–\ref{tab:stata_online}, simply replacing the textual branch with the \odaft{} encoder consistently yields higher Top-1 accuracy across diverse adaptation regimes. These findings align with the results observed in other settings, showcasing the broad generality of the proposed \odaft{} across different learning settings.

For MTA~\cite{zanella2024mta} and TPT~\cite{shu2022tpt},
\odaft{} improves the average Top-1 accuracy by \textbf{+1.7\%} and \textbf{+1.4\%}, respectively, across the 11 datasets.
The gains are especially pronounced for MTA on challenging, appearance-sensitive datasets such as EuroSAT (\textbf{+2.94\%}), DTD (\textbf{+2.11\%}), and Flowers102 (\textbf{+3.56\%}), where fine-grained categories exhibit strong visual overlap.
Despite the fact that both MTA and TPT adapt feature statistics dynamically at test time, the orthogonal textual space enforced by \odaft{} reduces prototype interference and stabilizes the prediction landscape, yielding consistent accuracy improvements across datasets.

StatA~\cite{stata2025cvpr} provides a more realistic and challenging evaluation through two protocols: \textit{batch-realistic} and \textit{online-realistic}. In the \textit{batch-realistic} setup (Table~\ref{tab:stata_offline}), where only a few classes are observed per batch, \odaft{} consistently improves performance across all effective-class regimes. Under the \textit{Very Low} $(1$–$4)$ regime with batch size $64$, it increases accuracy from $70.35$ to $71.21$ (\textbf{+0.86}), and in the \textit{Medium} $(5$–$25)$ range from $67.39$ to $68.62$ (\textbf{+1.23}). With larger batch size $1000$, the improvements remain positive uniformly across tasks, 
reaching \textbf{+1.09} points on average and up to \textbf{+1.0} in the \textit{Very High} $(50$–$100)$ regime. The stronger relative gain in low-class conditions highlights that \odaft{} is especially effective when class context is sparse and textual priors play a dominant role in adaptation. Yet, the continued improvement in higher regimes confirms its general robustness as more visual diversity is introduced.

In the \textit{online-realistic} variant (Table~\ref{tab:stata_online}), we follow the StatA protocol, where input streams are temporally correlated according to a Dirichlet distribution controlled by the correlation parameter~$\gamma$. Under this setup, \odaft{} again delivers consistent gains across all correlation levels. Under strong temporal correlation (\textit{High}, $\gamma{=}0.001$), performance rises from $69.45$ to $70.54$ (\textbf{+1.09}), and for the more challenging \textit{Separate} protocol, from $69.06$ to $70.53$ (\textbf{+1.47}). The effect is more pronounced at smaller $\gamma$ values, where the batch composition changes slowly, leading to correlation-induced drift in standard text embeddings. Orthogonal textual anchors mitigate this issue by maintaining angular separation between class directions, thereby preserving decision stability as the test stream evolves.

Overall, across MTA, TPT, and both StatA protocols, \odaft{} strengthens the adaptability of frozen vision–language models without modifying their optimization or inference. 
Consistent gains from \textbf{+1.5\%} on average for MTA/TPT to over \textbf{+1.6\%} in batch-realistic StatA show that enforcing orthogonality in the textual space yields a more disentangled, stable geometry. This in turn enhances the model’s ability to recalibrate confidently under distribution or temporal shifts, establishing \odaft{} as a robust and training-free augmentation to existing TTA frameworks.

\begin{table}[!htbp]
\centering
\scriptsize
\setlength{\tabcolsep}{6pt}
\renewcommand{\arraystretch}{1.05}
\caption{\textbf{Effect of Training-Free Orthogonalization via SVD (MTA).} 
Closed-form SVD whitening underperforms our learnable orthogonality, highlighting the importance of optimizing the text encoder to preserve semantic structure.}
\label{tab:tta_mta_svd}
\begin{tabular}{lc}
\toprule
\textbf{Method} & \textbf{Average (11 datasets)} \\
\midrule
MTA (baseline)       & 65.87 \\
+\,SVD (closed-form) & 61.23 \\
\rowcolor{RowGray}+\,\textbf{\odaft{} (ours)} & \textbf{67.53} \\
\bottomrule
\end{tabular}
\end{table}

\noindent\textbf{Training-Free Orthogonalization via SVD.}
We now evaluate the training-free variant that replaces our soft-penalty formulation with an SVD-whitened closed-form solution (Sec.~\ref{svd}). As shown in Table~\ref{tab:tta_mta_svd}, this closed-form approach 
is a suboptimal solution: MTA accuracy drops from $65.87$ to $61.2$, whereas our learnable \odaft{} improves it to $67.5$. SVD yields geometrically orthogonal but semantically distorted prototypes, treating all classes as equally unrelated. In contrast, \odaft{} preserves meaningful relationships while still reducing redundancy, which is 
crucial for downstream adaptation.

\noindent\textbf{Visual Alignment of \odaft{} Representations.}
Revisiting the feature geometry in Figure~\ref{fig:tsne_eurosat}, we examine how \odaft{} reshapes the EuroSAT embedding space. 
In the zero-shot CLIP space, semantically related classes (e.g., \textit{Annual Crop Land}, \textit{Pasture Land}) exhibit overlapping visual regions, and their corresponding textual prototypes $v_i$ lie close together despite distinct semantics.
After applying \odaft{}, the refined class embeddings $x_i$ shift towards the true visual centers of their respective image clusters, producing tighter and more separable groupings.
The average displacement between refined and zero-shot prototypes,
$
\frac{1}{K}\sum_{i=1}^{K}\|x_i - v_i\|_2,
$
is $0.23$ (median $0.15$), indicating moderate but consistent semantic realignment. 
Moreover, the average intra-class cosine dispersion decreases from $0.17$ to $0.10$ ($\sim40\%$ reduction), reflecting substantially sharper and less diffused class-conditional densities. 
This indicates a consistent realignment and sharper class densities, confirming that \odaft{} improves both semantic anchoring and geometric regularity of feature space.
\section{Conclusion}
\label{sec:conclusion}

We introduced \odaft, a simple text-only fine-tuning strategy that enhances VLMs by enforcing orthonormality among class-name embeddings.
Without using images or altering the vision backbone, \odaft{} reshapes the textual space to yield task-specific, semantically coherent prototypes that better capture class discriminability. 
Evaluations across zero-shot, few-shot and test-time adaptation settings show consistent gains on eleven benchmarks. 
Designed as a plug-and-play replacement for existing text encoders, \odaft{} demonstrates that structured geometric regularization of the textual representations alone can significantly strengthen downstream multimodal performance.
{
    \small
    \bibliographystyle{ieeenat_fullname}
    \bibliography{main}
}
\appendix
\clearpage
\setcounter{page}{1}
\maketitlesupplementary

\section{Datasets and Prompt Templates}

Following the standard practice in the literature
\cite{radford2021clip,zhou2022coop,zanella2024boosting}, 
we evaluate on 11 widely used recognition benchmarks covering a broad range of domains, label granularities and visual statistics.
Fine-grained object datasets include OxfordPets (``Pets'') \cite{parkhi2012pets}, StanfordCars (``Cars'') \cite{krause2013stanfordcars} and FGVCAircraft (``Aircraft''); 
texture and material recognition is evaluated on Describable Textures (DTD) \cite{cimpoi2014dtd} and Food101 (``Food'') \cite{bossard2014food101}; 
scene and remote–sensing understanding is assessed on SUN397 (``SUN'') \cite{xiao2010sun397} and EuroSAT \cite{helber2019eurosat}; 
and generic visual recognition is covered by Flowers102 (``Flowers'') \cite{nilsback2008flowers102}, Caltech101 (``Caltech'') \cite{fei2004caltech101}, UCF101 (``UCF'') \cite{soomro2012ucf101} and ImageNet \cite{deng2009imagenet}.  
For all datasets we follow the official or commonly adopted train/test splits and evaluation protocols used in CLIP and CoOp works.
We summarize the basic statistics and task descriptions for each benchmark in Table~\ref{tab:supp-datasets}.
Table~\ref{tab:supp-prompts} lists the three prompt templates used to train the text encoder with class names for each dataset.
These templates are used \emph{only} for \odaft{} pre-training stage; all baselines are evaluated with their respective default prompts.
Together, these details are intended to facilitate faithful reproduction of our experiments.

\begin{table}[!h]
\centering
\scriptsize
\setlength{\tabcolsep}{3.5pt}
\renewcommand{\arraystretch}{1.05}
\caption{\textbf{Additional information on the evaluation datasets.} 
We follow the standard class splits and test protocols used in CLIP/CoOp works.}
\label{tab:supp-datasets}
\begin{tabular}{l l r r}
\toprule
\textbf{Dataset} & \textbf{Other name} & \textbf{\#classes} & \textbf{\#test samples} \\
\midrule
SUN397        & SUN397      & 397  & 19{,}850 \\
FGVCAircraft  & Aircraft    & 100  & 3{,}333  \\
EuroSAT       & EuroSAT     & 10   & 8{,}100  \\
StanfordCars  & Cars        & 196  & 8{,}041  \\
Food101       & Food101     & 101  & 30{,}300 \\
OxfordPets    & Pets        & 37   & 3{,}669  \\
Flowers102    & Flowers102  & 102  & 2{,}463  \\
Caltech101    & Caltech101  & 101  & 2{,}465  \\
DTD           & DTD         & 47   & 1{,}692  \\
UCF101        & UCF101      & 101  & 3{,}783  \\
ImageNet      & ImageNet    & 1000 & 50{,}000 \\
\bottomrule
\end{tabular}
\end{table}
\begin{table}[!htbp]
\centering
\scriptsize
\setlength{\tabcolsep}{4pt}
\renewcommand{\arraystretch}{1.05}
\caption{\textbf{Prompt templates used for training the text encoder on each dataset.} 
Curly braces \{\} indicate the position where the class name is inserted.}
\label{tab:supp-prompts}
\begin{tabular}{l p{0.72\linewidth}}
\toprule
\textbf{Dataset} & \textbf{Prompt templates} \\
\midrule
ImageNet 
& ``a photo of a \{\}'', ``a picture of a \{\}'', ``an image of a \{\}'' \\[2pt]

SUN397 
& ``a photo of a \{\}'', ``an indoor scene of \{\}'', ``an outdoor scene of \{\}'' \\[2pt]

FGVCAircraft 
& ``a photo of a \{\}, a type of aircraft'', ``an in-flight \{\} aircraft'', ``a parked \{\} airplane'' \\[2pt]

EuroSAT 
& ``a satellite photo of \{\}'', ``an aerial photo of \{\}'', ``a high-resolution satellite image of \{\}'' \\[2pt]

StanfordCars 
& ``a photo of a \{\} car'', ``a showroom photo of a \{\}'', ``a street photo of \{\}'' \\[2pt]

Food101 
& ``a photo of \{\}'', ``a plated dish of \{\}'', ``a close-up of \{\}'' \\[2pt]

OxfordPets 
& ``a photo of a \{\}'', ``a portrait of a \{\}'', ``a close-up photo of a \{\}'' \\[2pt]

Flowers102 
& ``a close-up photo of a \{\} flower'', ``a macro photograph of \{\}'', ``a garden photograph of \{\}'' \\[2pt]

Caltech101 
& ``a photo of a \{\}'', ``a centered photo of \{\}'', ``a studio photo of \{\}'' \\[2pt]

DTD 
& ``\{\} texture'', ``a close-up of \{\} pattern'', ``a macro shot showing \{\} texture'' \\[2pt]

UCF101 
& ``a photo of a person doing \{\}'', ``an image of someone performing \{\}'', ``a video frame of \{\}'' \\
\bottomrule
\end{tabular}
\end{table}

\section{Additional Benchmark Protocols}

\paragraph{Zero-shot evaluation.}
For zero-shot evaluation, we replace the original CLIP/MetaCLIP text encoder by our orthogonal encoder and keep the vision backbone frozen. All methods are evaluated on the same 11 benchmarks described in the main paper (Pets, Cars, Aircraft, DTD, Food101, SUN397, EuroSAT, Flowers102, Caltech101, UCF101, ImageNet) using the standard class splits and test protocols from prior CLIP/CoOp works~\cite{radford2021clip,zhou2022coop,zhou2022cocoop}.

\paragraph{Few-shot protocols (CoOp and CLAP).}
In the few-shot setting, we follow the standard CoOp and CLAP training protocols. For each dataset and shot value $K$, we randomly sample $K$ labeled examples per class on the training split and train CoOp~\cite{zhou2022coop} (context-based prompt learning) or CLAP~\cite{clap24} (adapter-augmented tuning) on these support sets. Our method is integrated by initializing all textual prototypes in CoOp/CLAP with the orthogonal embeddings produced by \odaft{}, while keeping the vision backbone and all other hyperparameters identical to the original implementations. We report the mean accuracy over five random $K$-shot splits for each dataset.

\paragraph{Test-Time Adaptation: MTA and TPT.}
For the TTA regime, we adopt the original protocols of MTA and TPT.  
\textbf{MTA.} MTA~\cite{zanella2024mta} models the class scores as a balanced mixture of Gaussians in the CLIP embedding space and performs one-shot transductive adaptation on the entire test set, assuming all classes are present in the batch. We plug our orthogonal prototypes into the MTA initialization and keep all optimization hyperparameters (such as, number of EM iterations, batch size, temperature) fixed to those used in the official implementation.  \\
\textbf{TPT.} For TPT~\cite{shu2022tpt}, we perform test-time prompt tuning on each individual test image. Following the original paper, we generate multiple augmented views of the image, minimize the marginal entropy across views and use confidence selection to discard high-entropy (low-confidence) augmentations. In our integration, only the initial textual prototypes are changed to \odaft{} embeddings; the CLIP backbone, augmentation pipeline, number of views, and optimization schedule are kept identical to the original TPT setup.

\paragraph{Realistic test-time adaptation: StatA.}
For StatA, we strictly follow the realistic batch and online TTA scenarios introduced in~\cite{stata2025cvpr}.  
\textbf{Batch-realistic setting.} Each task corresponds to a batch of test samples with a limited number of effective classes $K_{\mathrm{eff}}$. We consider the six ranges defined in StatA: Very Low (1–4 classes), Low (2–10), Medium (5–25), High (25–50), Very High (50–100), and All (all classes present). For each dataset and scenario, 1{,}000 tasks are generated and accuracy is averaged over tasks, exactly as in~\cite{stata2025cvpr}.  
\textbf{Online-realistic setting.} In the streaming scenario, test data arrive as a sequence of correlated batches. Following StatA, we control the temporal correlation between batches using a Dirichlet distribution with concentration parameter $\gamma$ over class proportions: low, medium, and high correlation correspond to $\gamma \in \{0.1, 0.01, 0.001\}$, and the \textit{Separate} setting feeds classes sequentially~\cite{stata2025cvpr}. We adopt the same number of tasks (100 streams per configuration), batch size, and optimization schedule as StatA.  
In both batch-realistic and online-realistic regimes, we simply replace CLIP’s original textual prototypes by \odaft{} and leave all StatA hyperparameters unchanged, isolating the effect of our orthogonal text encoder on robustness across varying $K_{\mathrm{eff}}$ and Dirichlet-controlled correlations.

\begin{table*}[!h]
\centering
\scriptsize
\setlength{\tabcolsep}{5.2pt}
\renewcommand{\arraystretch}{1.05}
\caption{\textbf{Full per-dataset comparison of training-free SVD orthogonalization vs.\ \odaft{} under MTA.}  
Closed-form SVD whitening harms performance across most datasets, whereas our learnable orthogonality improves accuracy consistently.}
\label{tab:tta_mta_svd_full}
\begin{tabular}{lcccccccccccc}
\toprule
\textbf{Method} &
\textbf{Pets} &
\textbf{SUN397} &
\textbf{Aircraft} &
\textbf{DTD} &
\textbf{EuroSAT} &
\textbf{Cars} &
\textbf{Food101} &
\textbf{Flowers} &
\textbf{Caltech101} &
\textbf{UCF101} &
\textbf{ImageNet} &
\textbf{Avg} \\
\midrule
MTA (baseline)
& 88.24 & 66.67 & 25.32 & 45.90 & 45.36 & 68.47 & 84.95 & 68.26 & 94.21 & 68.11 & 69.11 & 65.87 \\

+\,SVD (closed-form)
& 80.40 & 65.40 & 20.40 & 46.20 & 40.90 & 60.20 & 82.80 & 57.10 & 92.10 & 63.80 & 64.20 & 61.23 \\

\rowcolor{RowGray}
+\,\textbf{\odaft{} (ours)}
& \textbf{89.10} & \textbf{67.80} & \textbf{27.00} & \textbf{48.01} & \textbf{48.30} & \textbf{68.93} & \textbf{85.70} & \textbf{71.82} & \textbf{94.90} & \textbf{71.30} & \textbf{70.01} & \textbf{67.53} \\
\bottomrule
\end{tabular}
\end{table*}

\subsection{Extended Analysis of SVD Orthogonalization (MTA)}

We expand the Table 6 of the main paper, in Table~\ref{tab:tta_mta_svd_full}, to report the per-dataset comparison of the training-free SVD variant against both the MTA baseline and our learnable \odaft{} encoder. Across almost all datasets, we see that enforcing hard orthogonality in closed form via SVD degrades performance, while \odaft{} yields systematic gains.

Compared to the baseline MTA, SVD whitening reduces accuracy on 10 out of 11 datasets, with drops of $4$--$5$ points on EuroSAT (\mbox{$45.36\% \rightarrow 40.90\%$}) and ImageNet (\mbox{$69.11\% \rightarrow 64.20\%$}), and over $11\%$ on Flowers (\mbox{$68.26\% \rightarrow 57.10\%$}). Even on datasets where the drop is smaller (e.g., DTD, UCF101), SVD never recovers the original baseline performance. This confirms that globally rotating the prototype matrix to an exactly orthogonal basis removes useful semantic anisotropy: classes that should remain moderately related (e.g., visually similar object or scene categories) are forced to be equally distant, which disrupts the alignment between visual and textual features learned during pre-training.

In contrast, \odaft{} improves over MTA on every dataset, with average accuracy increasing from $65.87\%$ to $67.53\%$. The gains are especially pronounced on challenging, fine-grained or texture-centric benchmarks such as Aircraft (\mbox{$25.32\% \rightarrow 27.00\%$}), EuroSAT (\mbox{$45.36\% \rightarrow 48.30\%$}), DTD (\mbox{$45.90\% \rightarrow 48.01\%$}), and Flowers (\mbox{$68.26 \rightarrow 71.82\%\%$}). These are precisely the regimes where prototype interference is most problematic: small angular overlaps between semantically related classes translate into large error rates. By softly encouraging orthogonality in the text encoder (rather than imposing it in one closed-form step), \odaft{} spreads these classes apart while still preserving their relative structure to the image features, leading to consistent improvements across all 11 datasets.

\begin{figure}[!htbp]
    \centering
    \includegraphics[width=0.78\linewidth]{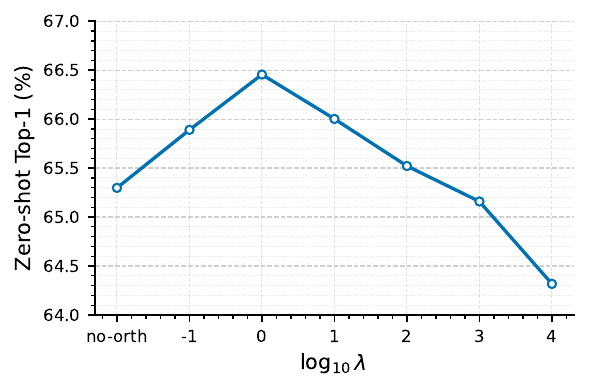}
    \caption{\textbf{Effect of the orthogonality penalty weight $\lambda$ on average zero-shot performance across 11 datasets.}
    We sweep $\log_{10}\lambda \in \{-1,0,1,2,3,4\}$, with ``no-orth'' corresponding to $\lambda=0$.
    Moderate regularization ($\lambda \approx 2$) yields the best performance, while overly large values degrade accuracy due to excessive hardening of class directions.}
    \label{fig:lambda_sweep}
\end{figure}

\section{Ablation on orthogonality penalty weight $\lambda$}

In Figure \ref{fig:lambda_sweep}, we examine the effect of the orthogonality penalty weight $\lambda$ from~\cref{eq:orth} of the main paper, on zero-shot accuracy averaged across all 11 datasets. The performance improves steadily as soft orthogonality is introduced. Accuracy rises from 65.4\% without orthogonality to 66.5\% at $\log_{10}\lambda=0$ (i.e., $\lambda=2$), marking the highest point in the curve. A slightly stronger penalty ($\log_{10}\lambda=1$, $\lambda=20$) maintains comparable performance (66.1\%).
Beyond this range, $\lambda \in [2, 20]$ , accuracy declines smoothly, e.g., 64.7\% at $\log_{10}\lambda=4$ ($\lambda=20000$), as excessive orthogonality over-constrains the geometry, forcing embeddings toward near-uniform orthogonality. 
This unimodal and smooth behavior highlights the robustness of \textsc{ORION}. 
Moderate soft orthogonality reliably enhances angular dispersion without collapsing meaningful class relationship, in contrast to rigid whitening (SVD).

\section{Comparison with prompt augmentation methods}

\begin{table*}[!h]
\centering
\scriptsize
\setlength{\tabcolsep}{3pt}
\renewcommand{\arraystretch}{1.0}
\caption{\textbf{Integrating ORION in prompt augmentation with CuPL (ViT-L/14).}
Replacing the baseline text encoder in CuPL with \textbf{\odaft} consistently improves zero-shot Top-1 accuracy across datasets.}
\label{tab:cupl_orion}
\begin{adjustbox}{max width=\textwidth}
\begin{tabular}{l*{11}{c}c}
\toprule
\textbf{Method} &
\textbf{Pets} &
\textbf{SUN397} &
\textbf{Aircraft} &
\textbf{DTD} &
\textbf{EuroSAT} &
\textbf{Cars} &
\textbf{Food101} &
\textbf{Flowers} &
\textbf{Caltech101} &
\textbf{UCF101} &
\textbf{ImageNet} &
\textbf{Avg} \\
\midrule
CuPL &
93.8 & 73.31 & 36.11 & 61.7 & 56.6 & 77.6 & 93.36 & 79.7 & 93.5 & 78.36 & 76.7 & 74.60 \\
\rowcolor{RowGray}
CuPL w/ ORION &
\bf94.2\gsub{+0.39} &
\bf73.69\gsub{+0.38} &
\bf36.6\gsub{+0.49} &
\bf61.85\gsub{+0.15} &
\bf58.67\gsub{+2.11} &
\bf78.1\gsub{+0.47} &
\bf93.7\gsub{+0.34} &
\bf79.9\gsub{+0.23} &
\bf95.9\gsub{+2.45} &
\bf79.41\gsub{+1.05} &
\bf77.21\gsub{+0.52} &
\bf75.38\gsub{+0.78} \\
\bottomrule
\end{tabular}
\end{adjustbox}
\end{table*}
\begin{table*}[!h]
\centering
\scriptsize
\setlength{\tabcolsep}{3pt}
\renewcommand{\arraystretch}{1.0}
\caption{\textbf{ORION complements zero-shot text refinement (ZLaP) (ViT-B/16).}
Integrating \textbf{\odaft} into ZLaP further improves performance, indicating that ORION enhances existing text-refinement pipelines.} 
\label{tab:zlap_orion}
\begin{adjustbox}{max width=\textwidth}
\begin{tabular}{l*{11}{c}c}
\toprule
\textbf{Method} &
\textbf{Pets} &
\textbf{SUN397} &
\textbf{Aircraft} &
\textbf{DTD} &
\textbf{EuroSAT} &
\textbf{Cars} &
\textbf{Food101} &
\textbf{Flowers} &
\textbf{Caltech101} &
\textbf{UCF101} &
\textbf{ImageNet} &
\textbf{Avg} \\
\midrule
ZLaP &
87.9 & 67.77 & 26.28 & 45.98 & 57.67 & 66.8 & 87.16 & 67.88 & 91.85 & 73.8 & 69.69 & 67.53 \\
\rowcolor{RowGray}
ZLaP w/ ORION &
88.4\gsub{+0.50} &
67.75\gsub{-0.02} &
25.2\gsub{-1.08} &
50.35\gsub{+4.37} &
59.64\gsub{+1.97} &
68.93\gsub{+2.13} &
87.1\gsub{-0.06} &
75.11\gsub{+7.23} &
93.91\gsub{+2.06} &
71.61\gsub{-2.19} &
70.01\gsub{+0.32} &
68.91\gsub{+1.38} \\
\bottomrule
\end{tabular}
\end{adjustbox}
\end{table*}

We now assess the complementary nature of \odaft{} on prompt augmentation methods such as CuPL~\cite{pratt2023does}. CuPL expands the textual search space by leveraging external LLMs to generate diverse class descriptions, thereby improving coverage of semantic variations. Since \odaft{} operates directly on the text encoder and refines the geometry of text prototypes, we initialize CuPL with \odaft{}'s refined text-encoder and prototypes.
Table~\ref{tab:cupl_orion} demonstrates that \odaft{} can be seamlessly integrated into prompt-augmentation pipeline to further boost performance, with notable accuracy gains on EuroSAT (56.6\% $\rightarrow$ 58.67\%, +2.11) and UCF101 (78.36 $\rightarrow$ 79.41\%, +1.05\%).

\section{Comparison to text refinement methods}

We further evaluate the compatibility of \odaft{} with zero-shot text refinement methods such as ZLaP\cite{kalantidis2024label}, which adapt textual representations using unlabeled test data. While ZLaP already improves alignment through transductive updates, ORION operates by refining the geometry of class prototypes in the text embedding space. As shown in Table~\ref{tab:zlap_orion}, combining the two yields consistent overall gains (+1.38\% on average), with notable improvements on Flowers (67.88 $\rightarrow$ 75.11, +7.23), DTD (45.98 $\rightarrow$ 50.35, +4.37), and Cars (66.8 $\rightarrow$ 68.93, +2.13). These gains suggest that ORION enhances class separability in regimes where fine-grained or texture-based distinctions are critical. Overall, the results confirm that ORION remains broadly complementary to existing zero-shot text refinement strategies.



\end{document}